\documentclass{article}

\usepackage{arxiv}

\usepackage[utf8]{inputenc} 
\usepackage[T1]{fontenc}    
\usepackage{hyperref}       
\usepackage{url}            
\usepackage{booktabs}       
\usepackage{amsfonts}       
\usepackage{nicefrac}       
\usepackage{microtype}      
\usepackage{lipsum}		
\usepackage{graphicx}
\usepackage[numbers]{natbib}
\usepackage{doi}
\usepackage{float}
\usepackage{booktabs}
\usepackage{array}
\usepackage{amsmath}

\title{\textbf{CNN-LSTM Hybrid Model for AI-Driven Prediction of COVID-19 Severity from Spike Sequences and Clinical Data}}

\date{} 					

\author{ \href{https://orcid.org/0000-0001-7826-9139}{\includegraphics[scale=0.06]{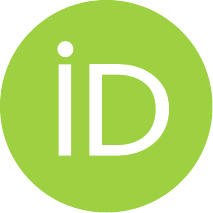}\hspace{1mm}Caio Cheohen}\thanks{Corresponding author} \\
	Universidade Federal do Rio de Janeiro\\
    Programa de Pós-graduação Multicêntrico em Ciências Fisiológicas \\
	NUPEM / UFRJ - Instituto de Biodiversidade e Sustentabilidade\\
		\texttt{caiocheohen@ufrj.br} \\
	\And
	\href{https://orcid.org/0000-0002-7417-1369}{\includegraphics[scale=0.06]{orcid.pdf}\hspace{1mm}Vinnícius Machado Schelk Gomes} \\
	Fundação Oswaldo Cruz (Fiocruz)\\
	Programa de Pós-Graduação Stricto sensu em Biologia Computacional e Sistemas \\
	\texttt{vinniciusgomes@aluno.fiocruz.br} \\
    \And
	\href{https://orcid.org/0000-0003-4844-7138}{\includegraphics[scale=0.06]{orcid.pdf}\hspace{1mm}Manuela Leal da Silva*} \\
	Universidade Federal do Rio de Janiero\\
	NUPEM / UFRJ - Instituto de Biodiversidade e Sustentabilidade \\
	\texttt{manuela@macae.ufrj.br} \\
}

\renewcommand{\headeright}{}
\renewcommand{\undertitle}{}
\renewcommand{\shorttitle}{\textit{CNN-lSTM Hybrid Model For AI-Driven Prediction Of COVID-19 Severity From Spike Sequences And Clinical Data} }

\hypersetup{
pdftitle={A template for the arxiv style},
pdfsubject={CNN-LSTM HYBRID MODEL FOR AI-DRIVEN PREDICTION OF COVID-19 SEVERITY FROM SPIKE SEQUENCES AND CLINICAL DATA},
pdfauthor={Caio Cheohen, Manuela Leal da Silva},
pdfkeywords={CNN-LSTM · Deep Learning · COVID-19 Severity · Machine Learning in Genomics · Clinical Outcome
Prediction},
}

\begin{document}
\maketitle
\begin{abstract}
\textbf{Background: }The COVID-19 pandemic, caused by SARS-CoV-2, highlighted the critical need for accurate prediction of disease severity to optimize healthcare resource allocation and patient management. The spike protein, which facilitates viral entry into host cells, exhibits high mutation rates, particularly in the receptor-binding domain (RBD), influencing viral pathogenicity. Artificial intelligence (AI) approaches, such as deep learning, offer promising solutions for leveraging genomic and clinical data to predict disease outcomes. \textbf{Objective: }This study aimed to develop a hybrid CNN-LSTM deep learning model to predict COVID-19 severity using spike protein sequences and associated clinical metadata from South American patients. \textbf{Methods: }We retrieved 9,570 spike protein sequences from the GISAID database, of which 3,467 met inclusion criteria after standardization. The dataset included 2,313 severe and 1,154 mild cases. A feature engineering pipeline extracted features from sequences, while demographic and clinical variables were one-hot encoded. A hybrid CNN-LSTM architecture was trained, combining CNN layers for local pattern extraction and an LSTM layer for long-term dependency modeling. \textbf{Results: }The model achieved an F1 score of \textbf{82.92\%}, ROC-AUC of \textbf{0.9084}, precision of \textbf{83.56\%}, and recall of \textbf{82.85\%}, demonstrating robust classification performance. Training stabilized at \~85\% accuracy with minimal overfitting. The most prevalent lineages (P.1, AY.99.2) and clades (GR, GK) aligned with regional epidemiological trends, suggesting potential associations between viral genetics and clinical outcomes.\textbf{Conclusion: }The CNN-LSTM hybrid model effectively predicted COVID-19 severity using spike protein sequences and clinical data, highlighting the utility of AI in genomic surveillance and precision public health. Despite dataset limitations, this approach provides a framework for early severity prediction in future outbreaks.
\end{abstract}


\section{Introduction}
\paragraph{} In December 2019, SARS-CoV-2 (Severe Acute Respiratory Syndrome Coronavirus 2) was discovered in Wuhan, China. The virus rapidly spread, causing the COVID-19 (Coronavirus Disease 2019) pandemic, which was associated with high mortality rates\cite{lu2020outbreak}. One of the most significant challenges posed by the pandemic is predicting the severity of infection in patients, especially in a scenario where high mutation rates are found in the SARS-CoV-2 spike protein\cite{qian2024predictive}\cite{de2021predicting}\cite{guruprasad2021human}\cite{muller2021sensitivity}. The spike protein is responsible for the interaction that leads to the attachment of the virus to human host cells and exhibits high mutation rates in the residues located in its receptor-binding domain (RBD). Even after the end of the pandemic, an accurate prediction of COVID-19 severity can help in early intervention, appropriate allocation of healthcare resources, genomic surveillance, and ultimately save lives. Aiming to solve this problem, machine learning algorithms can estimate the probability of disease progression using the amino acid residue sequences of the spike protein together with available clinical data\cite{liu2024predicting}.

To enable such analyses, open-access platforms for sharing genomic data have played a pivotal role throughout the pandemic\cite{carvalho2025empowering}\cite{khare2021gisaid}. The Global Initiative on Sharing All Influenza Data (GISAID) has served the scientific community as a repository for SARS-CoV-2 genomic sequences since the early stages of the outbreak\cite{khare2021gisaid}. GISAID provides raw viral genome data and associated metadata, such as collection date, location, and viral clade and lineage, which are essential for genomic surveillance and for developing predictive models of disease severity. GISAID has empowered researchers to track viral evolution, identify emerging variants, and train artificial intelligence models capable of anticipating clinical outcomes based on viral genetics\cite{khare2021gisaid}\cite{elbe2017data}\cite{perez2023innovative}.

Artificial intelligence (AI) has emerged as a powerful tool against COVID-19; in addition to diagnostic tools, AI has been used to predict COVID-19 infection severity\cite{chadaga2024explainable}. The scientific community has employed machine learning and deep learning approaches, such as Convolutional Neural Network (CNN) and Long Short-Term Memory (LSTM), to name a few\cite{lee2021severity}\cite{kim2025covid}\cite{muhammad2022cnn}\cite{sethi2024machine}.  A hybrid CNN-LSTM model combines the strengths of both CNN and LSTM to predict COVID-19 severity\cite{muhammad2022cnn}. This hybrid model can take clinical data such as age, gender, lineages, clade, and spike protein FASTA sequences as input. The CNN component of the model extracts relevant features from the data, while the LSTM component processes the temporal dynamics of the data\cite{kim2025covid}\cite{muhammad2022cnn}. Individually, CNN models are capable of extracting relevant local patterns from ordered sequences, such as amino acid sequences \cite{zhou2015predicting}\cite{li2019deep}\cite{min2017deep}. Proteins are formed by amino acid chains where local relations between residues, such as conserved motifs, functional domains, and structural interactions, strongly influence biological and functional protein properties\cite{chothia1986relation}\cite{jacobsen2023introduction}. While CNNs focus on spatial patterns, LSTMs are derived from Recurrent Neural Network (RNNs), designed to learn long-term dependencies, effectively addressing issues such as the vanishing gradient problem. They are widely used in tasks such as machine translation, time series forecasting, and anomaly detection\cite{sherstinsky2020fundamentals}\cite{ghojogh2023recurrent}. The use of a hybrid CNN-LSTM architecture typically involves an initial feature extraction step using a CNN, followed by temporal modeling with an LSTM \cite{sinha2025high}.

\paragraph{} In this study, we developed a deep learning forecasting model to predict COVID-19 severity, employing a hybrid CNN-LSTM architecture to explore the complementarity between two distinct types of sequential representations. The CNN was employed as a mechanism for local pattern extraction within the input features. The outputs of these convolutional layers are then passed to an LSTM layer, which functions as a temporal memory mechanism, allowing the model to capture long-range dependencies along the sequence. This is crucial in proteins, where residues that are distant in the linear chain may have interdependent functions or jointly contribute to the same functional structure or active site. The LSTM is particularly effective at recognizing patterns that depend on the context of the entire sequence, something a CNN alone would not be able to capture accurately. The integration of these two approaches enables the model to benefit both from the efficient local pattern extraction of CNNs and the modeling of global dependencies and long-term relationships provided by LSTMs, resulting in an expressive architecture, robust, and biologically coherent with the nature of the data.
\newpage
\section{Methods}
\label{sec:headings}

\subsection{\textbf{GISAID  Metadata Containing Patient Clinical Status}}

\paragraph{}Spike protein FASTA sequences and associated patient metadata derived from South American countries were retrieved from January to March 2023 at the GISAID database (\href{http://www.gisaid.org/}{http://www.gisaid.org}). For inclusion in this study, the sequences were required to meet the following criteria for the FASTA sequences of the SARS-CoV-2 S protein: a complete genome, human host, high coverage, and the presence of patient status and complete collection date in the metadata. According to GISAID, high coverage is defined as sequences containing less than 1\% undefined bases (NNNs), and insertions or deletions are only accepted if they have been verified by the submitter\cite{desouza_2021}. Patient metadata in GISAID often consists of non-standardized free-text entries, which may contain misspellings, abbreviations, and terms in various languages. To enable consistent downstream analysis, all metadata entries related to patient clinical condition were first mapped to a set of standardized "Status" labels, consolidating synonymous or variably spelled terms. These labels were subsequently grouped into broader clinical categories, namely "Mild" and "Severe", as summarized in \textbf{Table ~\ref{tab:classification_terms}}

\begin{table}[h!]
\centering
\caption{Mapping of Clinical Status Terms to Severity Classifications}
\label{tab:classification_terms}
\begin{tabular}{@{}cc@{}}
\toprule
\textbf{Classification} & \textbf{Clinical Status Terms} \\
\midrule
\textbf{Mild} & not hospitalized; alive/not hospitalized; Asymptomatic; Home; \\
             & not hospitalized; Not Hospitalized.; mild symptomatic; Mild; \\
             & Mild symptoms, not-hospitalized; No clinical signs; Not hospitalized \\
\addlinespace
\textbf{Severe} & DEAD; Dead, hospitalized; Death; deceased 14/8; deceased 20/8; \\
               & Decease; Deceased; Hospitalized (Intensive care unit); \\
               & Hospitalized, Live.; IC; Intensive Care; Intensive Care Unit; \\
               & severe symptomatic, required IC \\
\addlinespace
\textbf{Inconclusive} & ALIVE; Alive, hospitalized; Emergency Care; Hospitalized; \\
                     & Inpatient; Live; moderate symptomatic, hospita; Moderate \\
\bottomrule
\end{tabular}
\end{table}

\subsection{\textbf{Feature Extraction and Data Preprocessing}}
\paragraph{}To numerically represent protein sequences and associated metadata for supervised learning, we implemented a comprehensive feature engineering and preprocessing pipeline. We integrate global physicochemical descriptors, region-specific encodings, and demographic and clinical covariates, enabling the model to capture both general and domain-specific sequence characteristics.

\subsection{\textbf{Physicochemical and Structural Descriptors}}
\paragraph{} Each amino acid sequence was initially analyzed to extract global biochemical and structural features. The amino acid composition (AAC) was computed by normalizing the frequency of each residue over the sequence length. Additional global properties included sequence length, amino acid diversity, based on the number of unique residues found in the input sequence, mean hydrophobicity through Kyte–Doolittle scale\cite{kyte1982simple}, net charge at physiological pH (7.4) \cite{hopkins2025physiology}, and predicted secondary structure content (fraction of helix, strand, and coil), estimated using the \verb|ProteinAnalysis| module from Biopython\cite{cock2009biopython}. Polarity was also calculated using the Hopp–Woods scale\cite{hopp1981prediction}. by summing the polarity of each residue is weighted by its relative frequency in the sequence. Furthermore, to approximate the sequence's hydrogen bonding potential, the normalized frequency of residues capable of forming hydrogen bonds (Ser, Thr, Asn, Gln, His, and Tyr)\cite{zhou2019unraveling} was also computed.

\subsection{\textbf{Region-Specific Encoding: RBD-Focused Features}}
\paragraph{} To capture biologically relevant variations at the described interaction hotspots between the S protein and ACE2, we implemented a domain-aware encoding for the RBD, defined between residues 319 and 541\cite{verma2022mutation}.  Each residue in the sequence was represented by a 10-dimensional vector comprising normalized values of polarity, isoelectric point, hydrophobicity, and binary indicators for physicochemical classes (e.g., polar, charged, aromatic, aliphatic)\cite{kyte1982simple}\cite{hopkins2025physiology}\cite{cock2009biopython}\cite{hopp1981prediction}\cite{zhou2019unraveling}. A position-specific weighting scheme was applied\cite{heringa2002local}, to emphasize the functional importance of the RBD: residues within this region received a weight of 5, while all other positions were weighted by a factor of 1. Additionally, each residue was assigned a one-hot encoded secondary structure type — helix (H), strand (B), or coil (C).

\subsection{\textbf{Numerical Encoding and Sequence Padding}}
\paragraph{}All sequence-derived features were concatenated into a unified numeric vector and converted to NumPy arrays\cite{harris2020array}. To ensure compatibility with machine learning architectures, the vectors were padded with zeros to a fixed length of 3,013 elements, corresponding to the maximum observed sequence representation across the dataset.

\subsection{\textbf{Integration of Demographic and Clinical Variables}}

\paragraph{}Demographic (e.g., gender, age) and viral classification features (clade and lineage) were incorporated into the dataset using one-hot encoding via the  \verb|pandas.get_dummies|\cite{pandas2020pandas}. These encoded vectors were concatenated with the padded sequence features to form the final feature matrix (X), enabling the model to leverage both molecular and contextual information. The output variable (y) was defined by mapping the clinical severity status to binary labels: 'mild' was encoded as 1, and 'severe' as 0.

\subsection{\textbf{Feature Weighting Strategy}}
\paragraph{}To prevent overrepresentation of any single feature category (e.g., sequence or metadata), a uniform class weighting strategy was adopted\cite{santiago2020low}. Equal weights were assigned to both the FASTA-derived features and the clinical-demographic variables, promoting balanced model learning across heterogeneous input dimensions.

\subsection{\textbf{Model Training}}
\begin{figure}[h!]
    \centering
    \includegraphics[width=1\linewidth]{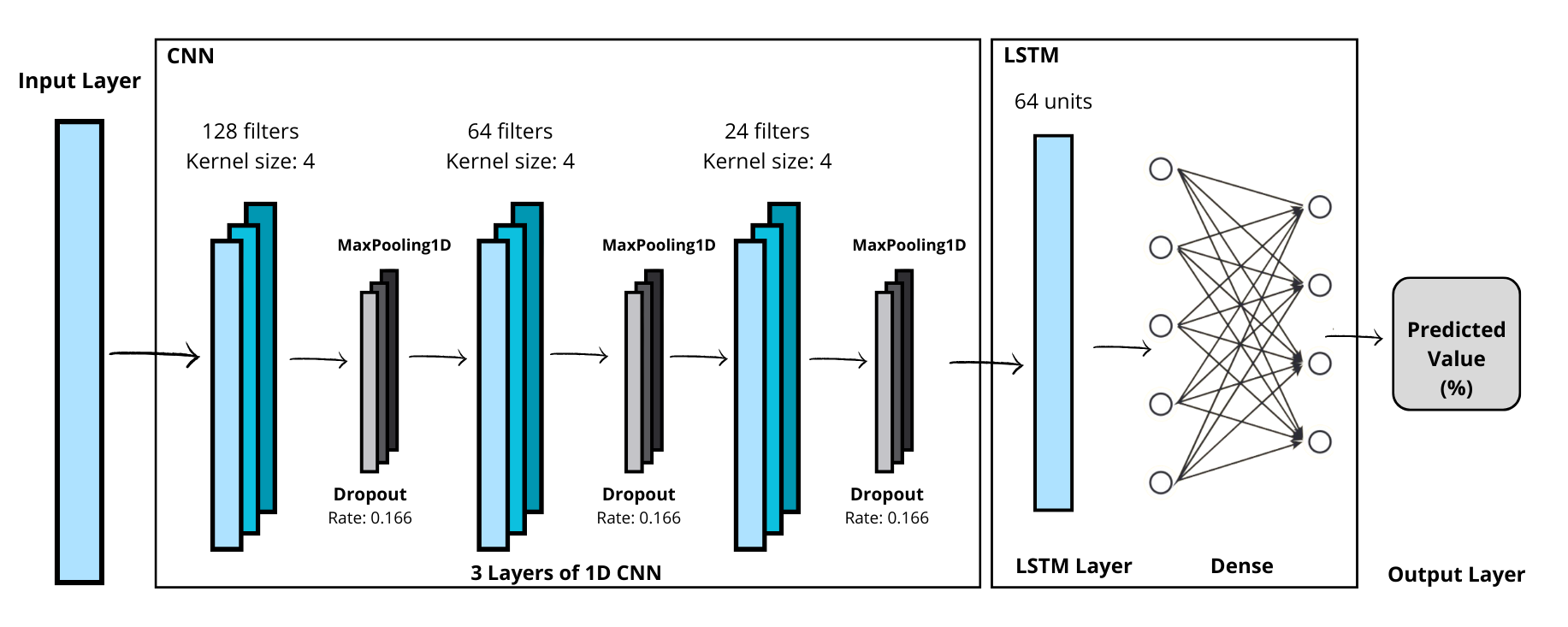}
    \caption{\textbf{CNN-LSTM Model Architecture --} Overview of the CNN-LSTM model used for prediction. The architecture includes three 1D convolutional layers with max pooling and dropout, followed by an LSTM layer and a dense layer to produce the final predicted value.
}
    \label{fig:CNN-LSTM-Model-Architecture}
\end{figure}

\paragraph{}The predictive model was developed using a hybrid approach based on CNN integrated with LSTM layers, implemented with the TensorFlow/Keras library\cite{abadi2015tensorflow}\cite{Keras}. Hyperparameter optimization was performed using the Optuna library\cite{akiba2019optuna},  which employs a Bayesian approach to identify the best parameter combinations. The final model includes a Conv1D layer with 128 filters (kernel size = 4), followed by MaxPooling1D (pool size = 2) and dropout (rate = 0.166). Two additional convolutional layers with 64 and 24 filters, respectively, are stacked, each followed by dropout. The final architecture comprised four convolutional layers followed by pooling layers, a 64-unit LSTM layer, multiple dense layers with ReLU activation, and a final layer with sigmoid activation for binary classification of the clinical outcome (e.g., "mild" vs. "severe")\cite{agarap2018deep} (\textbf{Figure} \ref{fig:CNN-LSTM-Model-Architecture}).

\newpage
\paragraph{}The deep learning architecture designed for this study comprises a total of 85,657 trainable parameters and no non-trainable parameters, indicating that all weights in the model are subject to optimization during training. A detailed summary of the model's architecture, including layer types, output shapes, and parameter counts, is provided in \textbf{Table ~\ref{tab:cnn_lstm_architecture}}

\begin{table}[h!]
\centering
\caption{Neural Network Architecture for Clinical Severity Prediction Based on Spike Protein Sequences}
\label{tab:cnn_lstm_architecture}
\begin{tabular}{@{}ccc@{}}
\toprule
\textbf{Layer (Type)} & \textbf{Output Shape} & \textbf{Parameters} \\
\midrule
Conv1D          & (None, 16727, 128) & 640 \\
MaxPooling1D    & (None, 8363, 128)  & 0 \\
Conv1D          & (None, 8360, 64)   & 32,832 \\
MaxPooling1D    & (None, 4180, 64)   & 0 \\
Conv1D          & (None, 4177, 64)   & 16,448 \\
MaxPooling1D    & (None, 2088, 64)   & 0 \\
Conv1D          & (None, 2085, 24)   & 6,168 \\
MaxPooling1D    & (None, 1042, 24)   & 0 \\
LSTM            & (None, 64)         & 22,784 \\
Dense           & (None, 64)         & 4,160 \\
Dropout         & (None, 64)         & 0 \\
Dense           & (None, 32)         & 2,080 \\
Dense           & (None, 16)         & 528 \\
Dense           & (None, 1)          & 17 \\
\midrule
\textbf{Total Parameters} & --- & \textbf{85,657} \\
\bottomrule
\end{tabular}
\end{table}

The dataset was split into 80\% training and 20\% testing subsets and balanced using the SMOTE (Synthetic Minority Over-sampling Technique) to mitigate class imbalance\cite{chawla2002smote}. Final training was conducted for 100 epochs using binary cross-entropy as the loss function and L2 regularization with a factor of 0.001 to prevent overfitting\cite{ng2004feature}. Evaluation metrics included F1 Score as the primary metric, along with Precision, Recall, ROC-AUC, Sensitivity, and Specificity\cite{yacouby2020probabilistic}.
\newpage
\section{Results}
\label{sec:headings}
\subsection{\textbf{GISAID  Metadata Containing Patient Clinical Status}}
\begin{figure}[h!]
    \centering
    \includegraphics[width=1\linewidth]{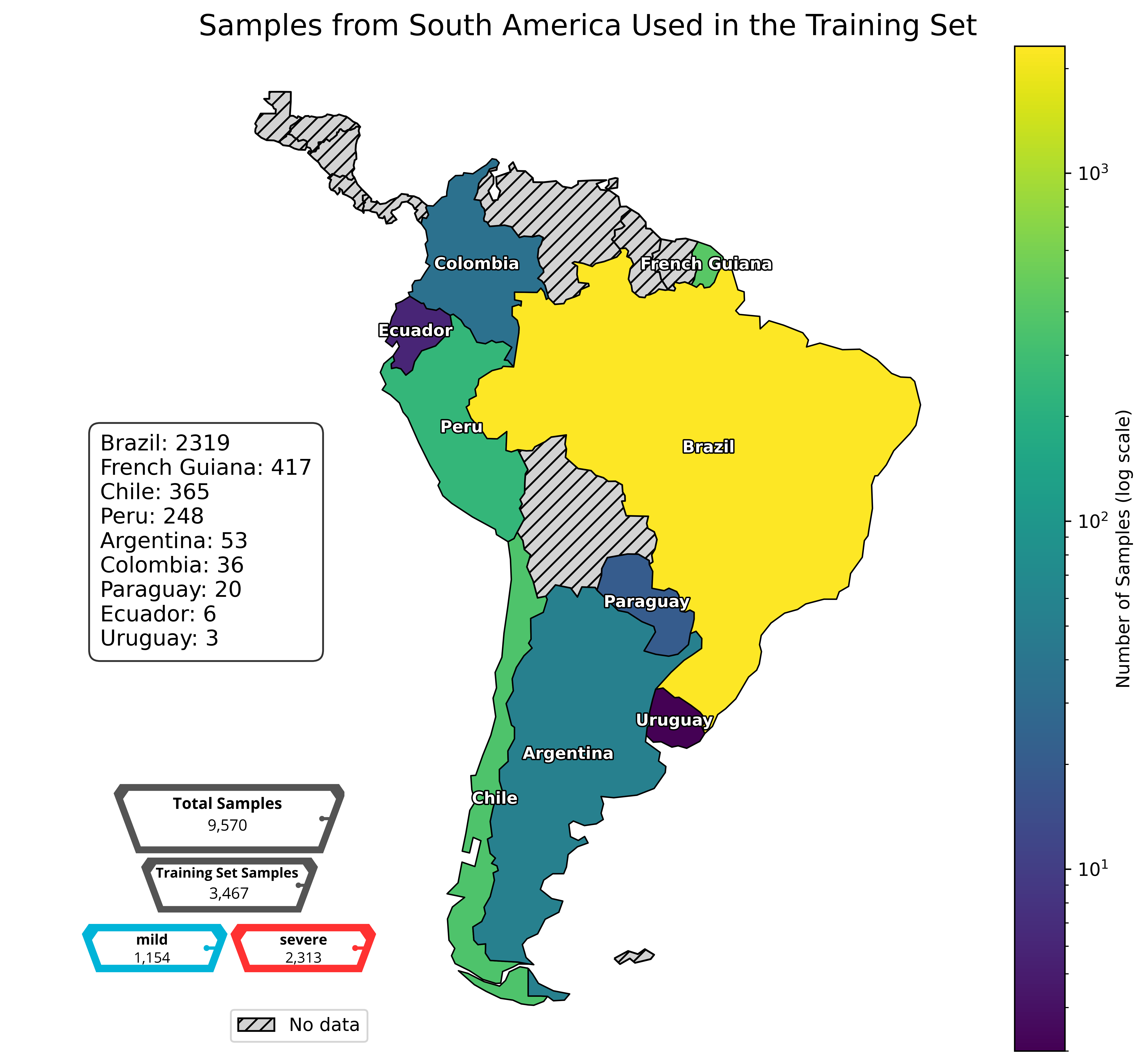}
    \caption{\textbf{SARS-CoV-2 Spike Protein Samples from South America --} Geographic distribution of samples used in the training set across South American countries. The color scale from green to yellow represents the number of samples on a logarithmic scale, highlighting variation between countries. Areas shaded with gray hatching indicate countries with no available data. The absolute number of samples per country is as follows: Brazil (2319), French Guiana (417), Chile (365), Argentina (53), Peru (248), Colombia (36), Paraguay (20), Ecuador (6), and Uruguay (3).}
    \label{fig:mapa_transparente}
\end{figure}

\paragraph{}A total of 9,570 spike protein sequences from South America were initially downloaded from GISAID, all meeting the predefined inclusion criteria. However, due to the heterogeneity and inconsistency in the patient metadata, only 3,467 samples could be retained after the standardization process described in the Methods section. Among these, 2,313 samples were classified as having a "Severe" clinical status, while 1,154 were categorized as "Mild" (\textbf{Figure} \ref{fig:mapa_transparente}).

\newpage
\paragraph{} Of the 3,467 samples included, 1,893 (54.6\%) were from male patients and 1,574 (45.4\%) from female patients. The mean age of the individuals in the dataset was 53.71 years overall, with males averaging 53.9 years and females 53.5 years. The lineage distribution revealed substantial diversity among the viral genomes. The most prevalent lineages were P.1 (n=961), AY.99.2 (n=363), AY.43.3 (n=245), B.1.1.33 (n=194), and B.1.1.28 (n=190) (\textbf{Figure} \ref{fig:lineage}), among others (see \textbf{Supplementary Table S1} for the complete list of identified lineages). Regarding phylogenetic clades, the majority of sequences belonged to clade GR (n=1,890) and clade GK (n=1,278). Other less frequent clades included GH (n=194), G (n=35), GRY (n=31), GRA (n=21), O (n=11), S (n=4), and GV (n=3).

\begin{figure}[h!]
    \centering
    \includegraphics[width=0.75\linewidth]{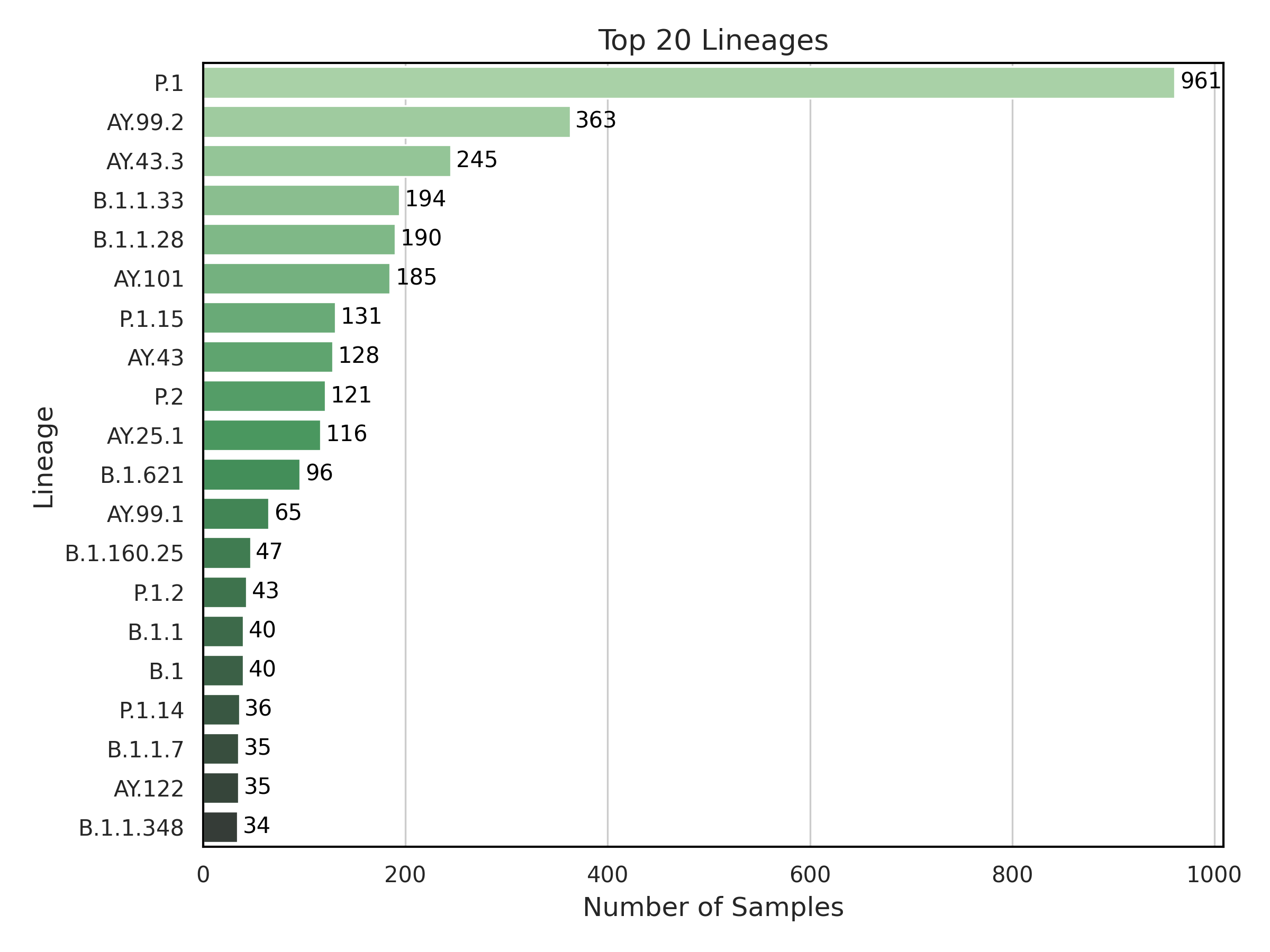}
    \caption{\textbf{Top 20 Lineages found in South America samples: }Retrieved lineages from January to March 2023 at the GISAID database after standardization process.}
    \label{fig:lineage}
\end{figure}
\subsection{\textbf{Model Training \& Performance Metrics}}

\paragraph{}A confusion matrix is an essential tool used to evaluate the performance of binary classifiers by summarizing how often predicted labels correspond to actual labels. For a binary classification problem, where the true labels can be either Positive or Negative\cite{Terven2025}. The components are defined as follows:

\begin{itemize}
    \item \textbf{True Positive (TP):} Instances correctly identified as positive.
    \item \textbf{True Negative (TN):} Instances correctly identified as negative.
    \item \textbf{False Positive (FP):} Occurs when the model incorrectly predicts a negative instance as positive.
    \item \textbf{False Negative (FN):} Occurs when the model incorrectly labels a positive instance as negative.
\end{itemize}

\paragraph{}During cross-validation, the optimal set of hyperparameters achieved a F1 score of 84.87\%. Using these parameters, the final model was trained for 100 epochs, and its performance in the test set is summarized in \textbf{Table ~\ref{tab:confusion_matrix}}.

\begin{table}[h!]
\centering
\caption{Confusion Matrix for Clinical Severity Classification}
\label{tab:confusion_matrix}
\begin{tabular}{@{}lcc@{}}
\toprule
 & \textbf{Predicted Negative} & \textbf{Predicted Positive} \\
\midrule
\textbf{Actual Negative} & 383 & 84 \\
\textbf{Actual Positive} & 37  & 190 \\
\bottomrule
\end{tabular}
\end{table}
\newpage

\paragraph{}The model’s predictive accuracy confirms that relevant sequence-level information for clinical severity classification can be captured from raw spike protein sequences, despite the absence of explicitly encoded biophysical properties (\textbf{Table ~\ref{tab:evaluation_metrics}}). Smaller convolutional kernels (size=4) enhanced local motif detection, while moderate dropout (0.166) prevented overfitting without sacrificing sensitivity. The LSTM layer (64 units) effectively captured temporal dependencies in sequence regions like the RBD domain.

\begin{table}[h!]
\centering
\caption{Evaluation Metrics After 100 Epochs of Training}
\label{tab:evaluation_metrics}
\begin{tabular}{@{}lc@{}}
\toprule
\textbf{Metric} & \textbf{Value} \\
\midrule
Precision Score & 0.8356 \\
Recall Score & 0.8285 \\
F1 Score & 0.8292 \\
ROC-AUC & 0.9084 \\
Sensitivity & 0.8370 \\
Specificity & 0.8201 \\
\bottomrule
\end{tabular}
\end{table}

\begin{figure}[h!]
    \centering
    \includegraphics[width=1\linewidth]{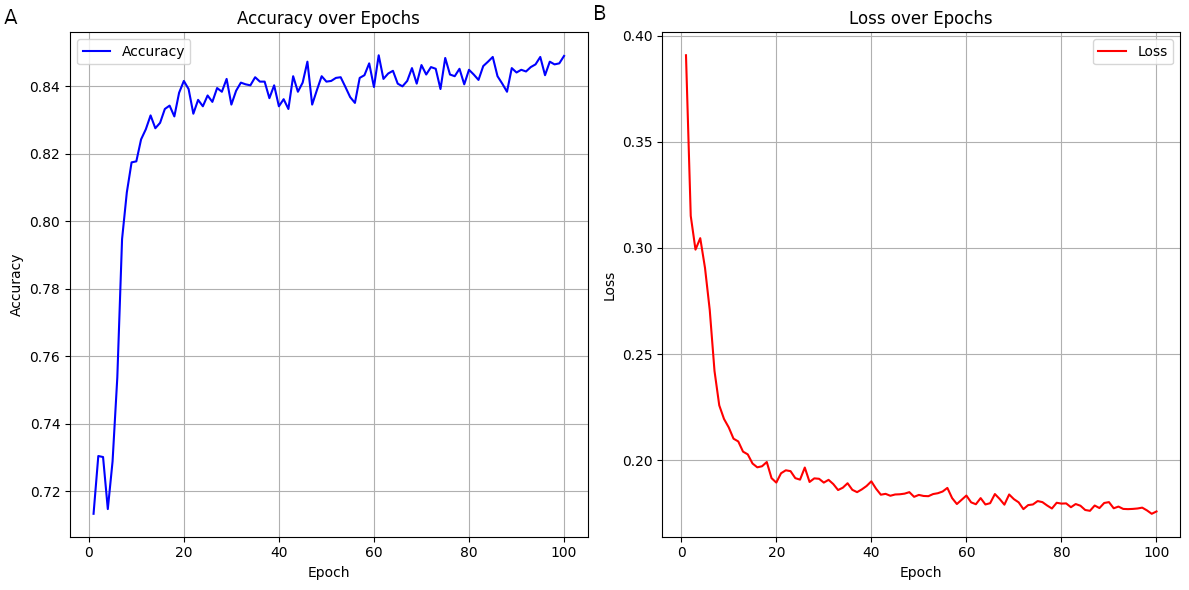}
    \caption{\textbf{Training Performance of the Deep Learning Model Across Epochs -- A: }The increase of accuracy over epochs. \textbf{B:} The decrease of loss over epochs.\textbf{ }}
    \label{fig:training-epochs}
\end{figure}

\paragraph{}The deep learning model showed steady performance gains during training. Accuracy rose from 70.6\% (loss = 0.387) at epoch 1 to 84.2\% (loss = 0.195) by epoch 20. From epochs 21 to 40, performance stabilized, with accuracy between 83.2\% and 84.8\%, and loss fluctuating around 0.19. In the final phase (epochs 41–60), accuracy peaked at 85.2\% (epoch 49), with loss remaining low (\( \approx 0.18 \)). These results suggest optimal learning around epoch 50, with minimal overfitting and strong generalization (\textbf{Figure} \ref{fig:training-epochs}).

\newpage

\section{Discussion}
\label{sec:Data Availability}
\paragraph{}The proposed model demonstrated consistent and strong performance across multiple evaluation metrics, with an F1 score of 82.92\% and ROC-AUC of 0.9084, confirming its effectiveness in binary classification of clinical severity. These results align with recent studies where models trained on SARS-CoV-2 genomic or spike protein sequences, sometimes combined with patient metadata, achieved similarly high predictive accuracy and discriminative power\cite{sokhansanj2022predicting}\cite{sokhansanj2022interpretable}.  The recall of 82.85\% further underscores the model’s potential for clinically actionable identification of severe cases. The high ROC-AUC suggests excellent discriminative capability, consistent with other studies employing deep learning for sequence-based classification tasks\cite{alipanahi2015predicting}\cite{zeng2016convolutional}. These results indicate that even without incorporating explicit biological or clinical features, neural networks can extract informative patterns from raw amino acid sequences, as previously observed in sequence-based variant effect predictors\cite{riesselman2018deep}\cite{bastani2021efficient}. 

\paragraph{}The performance trajectory during training, with accuracy increasing from 70.6\% to 85.2\% and stable loss values across epochs, indicates effective learning and minimal overfitting. The application of L2 regularization and a moderate dropout rate of 0.166 contributed to this stability, consistent with best practices in training regularized deep learning models\cite{soumare2021deep}\cite{varma2006bias}. The small discrepancy between cross-validation (F1 = 84.88\%) and final evaluation (F1 = 82.92\%) also supports the conclusion that the model generalizes well, albeit with minor overfitting.

\paragraph{}The predominance of the P.1 lineage and GR/GK clades among the sequences is consistent with known regional epidemiological trends in South America\cite{silva2021early}\cite{faria2021genomics}\cite{franceschi2021predominance}\cite{ribeirodias_2023}. These findings support the hypothesis that clade-specific mutations, particularly within the spike protein, may be associated with variations in clinical outcomes\cite{maurya2022sars}\cite{mendiola2022sars}\cite{mushebenge2023comprehensive}. Such associations reinforce previous evidence suggesting that certain mutations can influence viral behavior, including immune escape and disease severity\cite{faria2021genomics}\cite{harvey2021sars}\cite{planas2021reduced}\cite{zhou2021evidence}.

\paragraph{} Despite promising results, this study has several limitations. First, the dataset was imbalanced, necessitating the use of SMOTE for synthetic oversampling \cite{chawla2002smote}. While this technique helps mitigate bias in training, it may introduce an artificial signal that does not reflect true biological variance\cite{wang2021research}\cite{alkhawaldeh2023challenges}. \paragraph{}Another constraint was metadata heterogeneity, which led to a 64\% reduction in usable samples after standardization. This substantial data loss may have introduced biases in lineage distribution and reduced the overall representativeness of the dataset. Furthermore, the class imbalance (2,313 Severe vs. 1,154 Mild) may have inflated certain performance metrics, particularly specificity, although the strong recall suggests that clinically important cases were not overlooked.

\paragraph{}Notably, although our model was trained solely on spike protein sequences without incorporating structural or immunological covariates, it achieved strong predictive performance, emphasizing the rich information encoded in primary amino acid sequences. This aligns with recent advances demonstrating that deep learning models can accurately predict protein function and structure from sequence alone\cite{riesselman2018deep}\cite{bastani2021efficient}\cite{alley2019unified}.

\section{Conclusion}
\label{sec:Data Availability}

\paragraph{}This study demonstrates the feasibility of using deep learning models trained solely on spike protein sequences to predict COVID-19 clinical severity. Despite limitations in interpretability and data quality, the model achieved strong predictive performance, offering a valuable proof-of-concept for genomic-based clinical stratification. These results represent an important step toward the development of precision public health tools that leverage viral genomics to inform response strategies in future outbreaks.

\section{Acknowledgements}
\label{sec:Data Availability}

This work was supported by Fundação Carlos Chagas Filho de Amparo à Pesquisa do Estado do Rio de Janeiro (FAPERJ Processo SEI-260003/002152/2022 and FAPERJ/Proc 211.398/2019), Coordenação de Aperfeiçoamento de Pessoal de Nível Superior (CAPES) and the Conselho Nacional de Desenvolvimento Científico e Tecnológico / CNPq / Brasil.

\section{Data Avalability}
\label{sec:Data Availability}
All supplementary data supporting the findings of this study are provided in the Supplementary Materials. The source code and scripts used for model training and evaluation are publicly available at the following GitHub repository: \href{https://github.com/caiocheohen/AI-driven-COVID-health-stats-predictor}{https://github.com/caiocheohen/AI-driven-COVID-health-stats-predictor}.

\bibliographystyle{unsrt}
\bibliography{referencias}  






\end{document}